# Markov Equivalence Classes for Maximal Ancestral Graphs

R. Ayesha Ali
Thomas S. Richardson
Department of Statistics
University of Washington
BOX 354322
Seattle, WA 98195-4322
ayesha@stat.washington.edu

## Abstract

Ancestral graphs provide a class of graphs that can encode conditional independence relations that arise in directed acyclic graph (DAG) models with latent and selection variables, corresponding to marginalization and conditioning. However, for any ancestral graph, there may be several other graphs to which it is Markov equivalent. We introduce a simple representation of a Markov equivalence class of ancestral graphs, thereby facilitating the model search process for some given data. More specifically, we define a join operation on ancestral graphs which will associate a unique graph with an equivalence class. We also extend the separation criterion for ancestral graphs (which is an extension of d-separation) and provide a proof of the pairwise Markov property for joined ancestral graphs. Proving the pairwise Markov property is the first step towards developing a global Markov property for these graphs. The ultimate goal of this work is to obtain a full characterization of the structure of Markov equivalence classes for maximal ancestral graphs, thereby extending analogous results for DAGs given by Frydenberg (1990), Verma and Pearl (1991), Chickering (1995) and Andersson et al. (1997).

## 1  INTRODUCTION

A graphical Markov model is a set of distributions that can be described by a graph consisting of vertices and edges. The independence model associated with a graph is the set of conditional independence relations encoded by the graph. In this paper, we focus on the problem of learning causal structure. We suppose our observed data was generated by a process represented by a DAG with latent and selection variables. The causal interpretation of such a DAG is described by Spirtes et al. (1993), and Pearl (2000). There may be situations in which data collected from some process represented by a given data-generating process $\mathcal{D}$ is such that: i) measurement on some variables are unobserved (latent variables), and ii) some variables have been conditioned on (selection variables). One might think that in this case, though we may not be able to determine the influence of any hidden variables, we could just consider the observed variables and at least correctly represent the independence relations among them. Unfortunately, this is not always the case for DAG models because they are not closed under conditioning or marginalization. This point can be better understood through the following example.

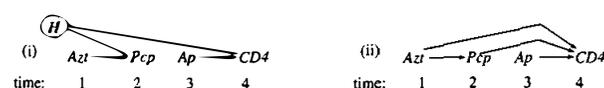

Figure 1: (i) A DAG with a latent variable $H$. (ii) A model search that does not include $H$ may add an extra edge from $Azt$ to $CD4$.

Consider the toy example given in Figure 1(i)*. $Azt$ is a drug given to AIDS patients to increase their CD4 counts. $Ap$ is a drug often given to AIDS patients to treat opportunistic infections. This graph pertains to the hypothetical experiment wherein subjects are randomized to $Azt$ at time 1 and $Ap$ at time 3, and then the outcome, $CD4$ count, is observed some time in the future. Suppose that there are side effects associated with $Azt$ such that some of the patients on $Azt$ develop the opportunistic infection $Pcp$, but $Azt$ has no effect on $CD4$ count. $H$ refers to a patient's underlying health status, which is not observed. A subject with poor health status may be more likely to develop $Pcp$ (observed at time 2), and she may also be more likely to have a low $CD4$ count. Note that temporal knowledge gives a total ordering on the variables.

---

*The example given in Figure 1 is a fictitious experiment based on an observational study analyzed by Hernán et al. (2000).

The DAG implies the following: $\langle Azt \perp\!\!\!\perp \{Ap, CD4\},$ $Ap \perp\!\!\!\perp \{Azt, Pcp\}\rangle$. In particular, note that $Azt$ is marginally independent of $CD4$. Given data generated by this DAG, a search over DAGs containing only the observed variables, and consistent with this time-ordering, would asymptotically find a DAG with an extra edge from $Azt$ to $CD4$ (see Figure 1(ii)). From such a search one could draw the incorrect conclusion that $Azt$ influences $CD4$ count. There is no DAG that can represent all of, and only, these independence relations using the observed variables alone. One approach to this problem would be to introduce latent variables into the model. However, introducing latent variables to a model may remove some of the desirable properties of the statistical distributions associated with the graph: these models may not be identifiable; the likelihood of the parameters for a specific model may be multi-modal; inference may be highly sensitive to the assumptions made about the unobserved variables; and the associated distributions may be difficult to characterize, in particular they may not form a curved exponential family. See Settimi and Smith (1999) and Geiger et al. (1999).

If detailed background knowledge is known about the process, then one might use a latent variable model, and exploit this information during the model search process. However, in the absence of background knowledge, we are in a dilemma: including latent variables explicitly can make modelling difficult, particularly when the structure of the graph is not known; *not* including hidden variables can potentially lead to misleading analyses (e.g. extra edges may be introduced to the graph). However, ancestral graphs are a class of graphs that, using only the observed variables, can encode the conditional independence relations given by any data-generating process that can be represented by a DAG with latent and selection variables. More precisely, it is shown in Richardson and Spirtes (2000) that if $\mathcal{D}$ is a DAG over the vertex set $V$ with latent variables $L$ and selection variables $S$, then there exists an ancestral graph $\mathcal{G}$ with vertex set $V\backslash(S \cup L)$ which is Markov equivalent to $\mathcal{D}$ on the $V\backslash(S \cup L)$ margin conditional on $S$. Furthermore, Richardson and Spirtes (2000) have shown that for any ancestral graph $\mathcal{G}$ (DAGs form a subset of ancestral graphs) with latent and selection variables, there are graphical operations corresponding to "marginalization" and "conditioning" such that the resulting graph represents the independence model obtained by taking the set of distributions represented by $\mathcal{G}$ and then integrating out the latent variables and conditioning on the selection variables. The resulting graph is itself an ancestral graph and represents the set of conditional independence relations holding among only the observed variables. Given the selection variables, the associated statistical models retain many of the desirable properties that are associated with DAG models.

However, as with DAG models, for any ancestral graph, there are potentially several other graphs that represent the same set of distributions. Such graphs are said to be *Markov equivalent*. Consequently, data cannot distinguish between Markov equivalent graphs. We define a join operation on ancestral graphs which associates a unique graph with an equivalence class. We also extend the separation criterion (See Definition 2.2) for ancestral graphs (which is an extension of d-separation) and provide an outline of the proof of the pairwise Markov property for joined ancestral graphs. Andersson et al. (1997) showed that the graph resulting from joining a Markov equivalence class of DAGs is a chain graph. They also characterized the structure of this chain graph and showed that it is Markov equivalent to the original DAGs in the equivalence class. The pairwise Markov property for joining DAGs follows from their finding. Partial characterizations of Markov equivalence classes for ancestral graphs have been obtained using POIPGs and PAGs by Richardson and Spirtes (2002) and Spirtes et al. (1993). A key difference between these authors' works and the present investigation is that the representation given here is guaranteed to include all arrowheads common to every graph in the equivalence class, whereas this is not true in the previous work. In other words, the representation here is guaranteed to be *complete* with respect to arrowheads (see Meek (1995)). The graphs described here are analogous to the essential graph for DAGs (Andersson et al. (1997)), while previous representations have been analogous to Patterns (Verma and Pearl (1991)).

Section 2 provides some basic definitions; Section 3 starts to characterize various aspects of joined graphs with respect to minimal inducing paths; Section 4 outlines the proof that the joined graph formed by joining Markov equivalent maximal ancestral graphs is itself maximal; and finally, Section 5 outlines areas for future research.

## 2 BASIC DEFINITIONS

### 2.1 VERTEX RELATIONS

If there is an edge between $\alpha$ and $\beta$ in the graph $\mathcal{G}$, then $\alpha$ is *adjacent* to (sometimes referred to as *"an adjacency of"*) $\beta$ and vice versa.

If $\alpha$ and $\beta$ are vertices in a graph $\mathcal{G}$ such that $\alpha \leftrightarrow \beta$, then $\alpha$ is a *spouse* of $\beta$ and vice versa.

If $\alpha$ and $\beta$ are vertices in a graph $\mathcal{G}$ such that $\alpha \to \beta$, then $\alpha$ is a *parent* of $\beta$, and $\beta$ is a *child* of $\alpha$.

If there is a directed path from $\alpha$ to $\beta$ (i.e. $\alpha \to \to \ldots \to \beta$) or $\alpha = \beta$, then $\alpha$ is an *ancestor* of $\beta$, and $\beta$ is a *descendant* of $\alpha$. Also, this directed path from $\alpha$ to $\beta$ is called an *ancestral path*.

### 2.2 ANCESTRAL GRAPHS

The basic motivation for developing ancestral graphs is to enable one to focus on the independence structure over the observed variables that results from the presence of latent variables without explicitly including latent variables in the model. Permitting bi-directed ($\leftrightarrow$) edges in the graph allows one to graphically represent the existence of an unobserved common cause of observed variables. For Figure 1(i) this corresponds to removing $H$ from the graph and adding a bi-directed edge between $Pcp$ and $CD4$. Undirected edges ($-$) are also introduced to represent unobserved selection variables that have been conditioned on rather than marginalized over. However, interpreting ancestral graphs is not so straightforward. Richardson and Spirtes (2002) provides a detailed discussion on the interpretation of edges in an ancestral graph. Further details of the basic definitions and concepts presented here can also be found in Richardson and Spirtes (2000).

**Definition 2.1** *A graph, which may contain undirected ($-$), directed edges ($\to$) and bi-directed edges ($\leftrightarrow$) is ancestral if:*
*(a) there are no directed cycles;*
*(b) whenever an edge $x \leftrightarrow y$ is in the graph, then $x$ is not an ancestor of $y$, (and vice versa);*
*(c) if there is an undirected edge $x - y$ then $x$ and $y$ have no spouses or parents.*

Conditions *(a)* and *(b)* may be summarized by saying that if $x$ and $y$ are joined by an edge and there is an arrowhead at $x$, then $x$ is *not* an ancestor of $y$; this is the motivation for the term 'ancestral'. Note that by *(c)*, the configurations $\to \gamma -$ and $\leftrightarrow \gamma -$ never occur in an ancestral graph.

A natural extension of Pearl's d-separation criterion may be applied to ancestral graphs. For ancestral graphs, a non-endpoint vertex $v$ on a path is said to be a *collider* if two arrowheads meet at $v$, i.e. $\to v \leftarrow$, $\leftrightarrow v \leftrightarrow$, $\leftrightarrow v \leftarrow$ or $\to v \leftrightarrow$; all other non-endpoint vertices on a path are *non-colliders*, i.e. $- v -$, $- v \to$, $\to v \to$, $\leftarrow v \to$.

**Definition 2.2** *In an ancestral graph, a path $\pi$ between $\alpha$ and $\beta$ is said to be m-connecting given $Z$ if the following hold:*

*(i) No non-collider on $\pi$ is in $Z$;*
*(ii) Every collider on $\pi$ is an ancestor of a vertex in $Z$.*

*Two vertices $\alpha$ and $\beta$ are said to be m-separated given $Z$ if there is no path m-connecting $\alpha$ and $\beta$ given $Z$.*

Definition 2.2 is an extension of the original definition of d-separation for DAGs in that the notions of 'collider' and 'non-collider' now include bi-directed and undirected edges. Since m-separation characterizes the independence relations in an underlying probability distribution compatible with a graph, tests of m-separation can be used to determine when graphs are Markov equivalent to each other.

**Definition 2.3** *Two graphs $\mathcal{G}_1$ and $\mathcal{G}_2$ are said to be Markov equivalent if for all disjoint sets $A, B, Z$ (where $Z$ may be empty), $A$ and $B$ are m-separated given $Z$ in $\mathcal{G}_1$ if and only if $A$ and $B$ are m-separated given $Z$ in $\mathcal{G}_2$.*

Independence models described by DAGs satisfy pairwise Markov properties such that every missing edge corresponds to a conditional independence relation. In general, this property does not apply to ancestral graphs. For example, there is no set which m-separates $\gamma$ and $\delta$ in the graph in Figure 2(a), which motivates the following definition:

**Definition 2.4** *An ancestral graph $\mathcal{G}$ is said to be "maximal" if, for every pair of non-adjacent vertices $\alpha, \beta$ there exists a set $Z (\alpha, \beta \notin Z)$, such that $\alpha$ and $\beta$ are m-separated conditional on $Z$.*

These graphs are termed *maximal* in the sense that no additional edge may be added to the graph without changing the associated independence model. It has been shown in Richardson and Spirtes (2000) that if an ancestral graph is not maximal, then there exists at least one pair of non-adjacent vertices $\{\alpha, \beta\}$, for which there is an "inducing path" between $\alpha$ and $\beta$ where:

**Definition 2.5** *An inducing path $\pi$ is a path in an ancestral graph such that each non-endpoint vertex is a collider, and an ancestor of at least one of the endpoints.*

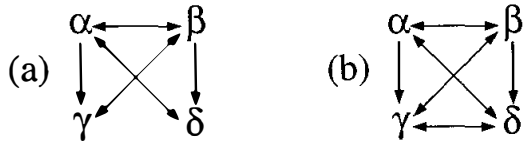

**Figure 2:** (a) The path $\{\gamma, \beta, \alpha, \delta\}$ is an example of an inducing path in an ancestral graph. (b) A maximal ancestral graph Markov equivalent to (a).

Figure 2(a) shows an example of a non-maximal ancestral graph. By adding a bi-directed edge between $\gamma$ and $\delta$, the graph can be made maximal, as shown in Figure 2(b).

**Definition 2.6** *Suppose $\langle \alpha, \beta, \delta \rangle$ are vertices in a graph such that $\alpha$ and $\beta$ are adjacent, and $\beta$ and $\delta$ are adjacent. If $\alpha$ and $\delta$ are also adjacent, then $\langle \alpha, \beta, \delta \rangle$ is "shielded". Otherwise, if $\alpha$ and $\delta$ are not adjacent, then $\langle \alpha, \beta, \delta \rangle$ is "unshielded".*

One of the key differences between DAGs and ancestral graphs is that there are some shielded colliders in ancestral graphs $\mathcal{G}$ that must be present in any other ancestral graph Markov equivalent to $\mathcal{G}$; considering shielded colliders is not important in determining Markov equivalence for DAGs. *Discriminating paths* are useful for identifying which shielded colliders (and non-colliders) are required for ancestral graphs to be Markov equivalent:

**Definition 2.7** $U = \langle x, q_1, q_2, \ldots, q_p, \beta, y \rangle$ *is a discriminating path for $\beta$ in an ancestral graph $\mathcal{G}$ if and only if:*

*(i) $U$ is a path between $x$ and $y$ with at least three edges,*
*(ii) $U$ contains $\beta, \beta \neq x, \beta \neq y$,*
*(iii) $\beta$ is adjacent to $y$ on $U$, $x$ is not adjacent to $y$, and*
*(iv) For every vertex $q_i, 1 \leq i \leq p$ on $U$, excluding $x, y$, and $\beta$, $q_i$ is a collider on $U$ and $q_i$ is a parent of $y$.*

Given a set $Z$, if $Z$ does not contain all $q_i, 1 \leq i \leq p$, then the path $\langle x, q_1, \ldots, q_j, y \rangle$ is m-connecting where $q_j \notin Z$ and $q_i \in Z$ for all $i < j$. If $Z$ contains $\{q_1, \ldots, q_p\}$ and $\beta$ is a collider on the path $U$ in the graph $\mathcal{G}$, then $\beta \notin Z$ if $Z$ m-separates $x$ and $y$. Consequently, in any graph Markov equivalent to $\mathcal{G}$ containing the discriminating path $U$, $\beta$ is also a collider on $U$. Similarly, if $\beta$ is a non-collider on the path $U$ then $\beta$ is a member of any set that m-separates $x$ and $y$, and $\beta$ is a non-collider on $U$ in any graph Markov equivalent to $\mathcal{G}$ containing $U$. In other words, $\beta$ is "discriminated" to be either a collider or a non-collider on the path $U$ in any graph Markov equivalent to $\mathcal{G}$ in which $U$ forms a discriminating path, even though it is shielded. The paths $\langle x, q, \beta, y \rangle$ in $\mathcal{G}_1$ and $\mathcal{G}_2$ from Figure 4 are examples of discriminating paths for $\beta$. Note that if $\beta$ is a non-collider on $U$, then $\beta \to y$ in $\mathcal{G}$.

**Definition 2.8** *A "collider path" in an ancestral graph $\mathcal{G}$ is a path such that every vertex, except the endpoints, is a collider on that path.*

From the definition of a discriminating path, the subpath of $U$ from $x$ to $\beta$ forms a *collider path*. So referring to $\mathcal{G}_1$ and $\mathcal{G}_2$ in Figure 4, the path $\langle x, q, \beta \rangle$ is a collider path (and in fact, in these examples, $\langle x, q, \beta, y \rangle$ forms a collider path too).

### 2.3 CHARACTERIZATION OF MARKOV EQUIVALENCE

Spirtes and Richardson (1997) proved the following result:

**Theorem 2.1** *(Markov Equivalence) Two maximal ancestral graphs $\mathcal{G}_1$ and $\mathcal{G}_2$ are Markov equivalent if and only if:*

*(i) $\mathcal{G}_1$ and $\mathcal{G}_2$ have the same adjacencies;*
*(ii) $\mathcal{G}_1$ and $\mathcal{G}_2$ have the same unshielded colliders; and*
*(iii) If $U$ forms a discriminating path for $\beta$ in $\mathcal{G}_1$ and $\mathcal{G}_2$, then $\beta$ is a collider in $\mathcal{G}_1$ if and only if it is a collider in $\mathcal{G}_2$.*

### 2.4 JOINED GRAPHS

Here we define the join operation as a method of identifying the features common to a set of Markov equivalent ancestral graphs. By definition, a set of Markov equivalent maximal ancestral graphs are required to have the same vertex set and adjacencies. The join operation can be thought of as an AND operation on the "arrowheads" of the set of Markov equivalent ancestral graphs being joined, and an OR operation on the "tails" of these graphs.

**Definition 2.9** *Let $\mathcal{G}_1, \mathcal{G}_2, \ldots, \mathcal{G}_n$ be graphs with the same adjacencies. A joined graph, $\mathcal{H}$ is any graph constructed in the following way:*

*(i) $\mathcal{H}$ has the same adjacencies as $\mathcal{G}_1, \mathcal{G}_2, \ldots, \mathcal{G}_n$,*
*(ii) For all adjacent $\alpha$ and $\beta$, add an arrowhead at $\beta$ on the $\{\alpha, \beta\}$ edge if and only if there is an arrowhead at $\beta$ on the $\{\alpha, \beta\}$ edge in all $\mathcal{G}_1, \mathcal{G}_2, \ldots, \mathcal{G}_n$.*

In general we will let $\mathcal{H}$ refer to a joined graph formed by joining any number of Markov equivalent maximal ancestral graphs. We will also generically refer to these maximal ancestral graphs as $\mathcal{G}$.

Figure 3 provides an example of a joined graph. Note that since there are arrowheads that meet the undi-rected edge $x - w$ in the joined graph, $\mathcal{H}$ is not ancestral as it violates condition (c) of Definition 2.1. Figure 4 shows another example of two Markov equivalent graphs being joined. Here, $\mathcal{H}$ is itself a member of the equivalence class of ancestral graphs.

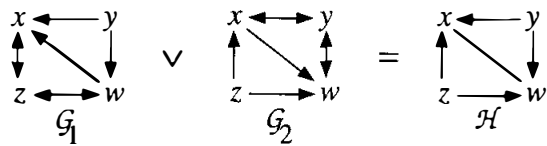

Figure 3: An example of joining two Markov equivalent ancestral graphs in which the joined graph is not ancestral.

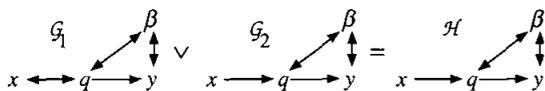

Figure 4: An example of joining two Markov equivalent ancestral graphs in which the joined graph is itself a member of the equivalence class.

Richardson and Spirtes (2000) showed that for every non-maximal ancestral graph $\mathcal{G}$, there exists a unique maximal ancestral graph which is formed by adding appropriate bi-directed ($\leftrightarrow$) edges to $\mathcal{G}$ (see Figure 2). Hence we restrict our attention to joining sets of Markov equivalent maximal ancestral graphs in the remainder of this paper. Ideally, any representation of an equivalence class of ancestral graphs would encode the same independence model encoded by all the ancestral graphs in the equivalence class.

We use the following notation for endpoints in either an ancestral graph or a joined graph:

1. "$\alpha - ?\beta$" is used to denote that there is a tail at $\alpha$ in the graph, on the edge between $\alpha$ and $\beta$, and that there may be a tail or an arrowhead at the $\beta$ end of this edge.
2. "$\alpha \leftarrow ?\beta$" is used to denote that there is an arrowhead at $\alpha$, and either an arrowhead or a tail at $\beta$ on the edge between $\alpha$ and $\beta$.
3. "$\alpha? - ?\beta$" is used to denote that there could be an arrowhead or tail at either end of the $\langle \alpha, \beta \rangle$ edge.

Note that the above notation is merely a shorthand since we only consider graphs with edges that are directed, bi-directed and undirected. By joining maximal ancestral graphs as outlined in Definition 2.9, the resulting joined graph $\mathcal{H}$ is not ancestral in general, see Figure 3. Given that undirected edges can meet arrowheads in joined graphs, what is the equivalent of a d-connecting path for joined graphs? Here we define a *j-connecting* path for joined graphs.

**Definition 2.10** *A path between $\alpha$ and $\beta$ in a joined graph $\mathcal{H}$ is said to be "j-connecting given a set $Z$" ($Z$ disjoint from $\{\alpha, \beta\}$ and possibly empty) if:*

(i) *Every non-collider ($? - \gamma - ?, ? \to \gamma \to, \leftarrow \gamma \leftarrow ?$) on the path is not in $Z$,*

(ii) *Every collider ($? \to \gamma \leftarrow ?$) on the path is an ancestor of $Z$, and*

(iii) *No arrowheads meet undirected edges.*

*If there is no path that j-connects $\alpha$ and $\beta$ given $Z$, then $\alpha$ and $\beta$ are "j-separated given $Z$".*

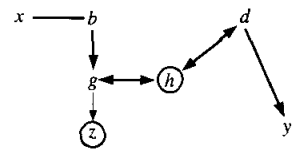

Figure 5: An example of a j-connecting path in a joined graph: $x$ and $y$ are j-connecting given $Z = \{z, h\}$.

Note that this definition is a natural extension of m-connection for ancestral graphs (and Pearl's d-connection for DAGs), with the qualifier that undirected edges meeting arrowheads form neither colliders nor non-colliders and a path containing such a vertex is never j-connecting. If we look back at the joined graph shown in Figure 3, we see that $\mathcal{H}$ encodes the same set of independence relations that the two ancestral graphs that gave rise to $\mathcal{H}$ encode, namely $y \perp\!\!\!\perp z$, because there are no j-connecting paths between $y$ and $z$ in $\mathcal{H}$ (the path $z \to x - w \leftarrow y$ is not j-connecting). Figure 5 shows another example of a j-connecting path. Here, some vertices in $Z$ are descendants of colliders on the path between $x$ and $y$.

The definitions of discriminating paths and inducing paths for joined graphs remain the same as for ancestral graphs. Here we extend the concept of maximality to joined graphs and in Section 4 we show that the graph $\mathcal{H}$ formed by joining Markov equivalent maximal ancestral graphs is itself maximal.

**Definition 2.11** *A joined graph $\mathcal{H}$ is said to be "maximal" if, for every pair of non-adjacent vertices $\alpha, \beta$ there exists a set $Z(\alpha, \beta \notin Z)$, such that $\alpha$ and $\beta$ are j-separated conditional on $Z$.*

The concept of maximality for joined graphs is analogous to that for ancestral graphs in that a maximal joined graph is a joined graph, $\mathcal{H}$, such that no more edges can be added to $\mathcal{H}$ without changing the set of independence relations encoded by $\mathcal{H}$ via j-separation.

# 3 CHARACTERIZING THE JOINED GRAPH

To date, no full characterization of joined graphs is readily available. This section presents structural inferences that can be made about joined graphs. For instance, as with ancestral graphs, the configurations "$\dashrightarrow \gamma -$" and "$\leftrightarrow \gamma -$" do not occur in joined graphs. We also conjecture that the graph resulting from joining an entire equivalence class of ancestral graphs can be more constrained than that obtained by joining only a few members of an equivalence class.

If an edge is oriented the same way in all graphs $\mathcal{G}$ that were joined to form $\mathcal{H}$, then that edge is said to be "*real*" in $\mathcal{H}$. By virtue of the join operation, it is possible to infer the presence of arrowheads and tails in joined graphs under certain circumstances. The following lemmas describe some of these situations.

**Lemma 3.1** *All bi-directed edges in a joined graph are real. Furthermore, if $\alpha? \to \beta - \gamma$ in $\mathcal{H}$, then the $\beta - \gamma$ edge is not real.*

**Proof:** By the definition of the join operation, an arrowhead appears at a vertex in the joined graph $\mathcal{H}$ if and only if there is an arrowhead at that vertex in all ancestral graphs that gave rise to $\mathcal{H}$. Also, no ancestral graph contains undirected edges meeting arrowheads, so if an undirected edge meets an arrowhead in a joined graph (using the example given in the proposition) then there is at least one ancestral graph that gave rise to $\mathcal{H}$ with an arrowhead at $\beta$ on the $\beta - \gamma$ edge, i.e. the $\beta - \gamma$ edge is not real.

**Lemma 3.2** *In a joined graph $\mathcal{H}$, formed by joining maximal ancestral graphs, if $\gamma \to \beta? - ?\delta? \to \gamma$ occurs and $\gamma \to \beta$ is real, then $\beta \leftarrow ?\delta$ also occurs in $\mathcal{H}$.*

**Proof:** There cannot be a tail at $\beta$ on the $\{\beta, \delta\}$ edge in any ancestral graph that gave rise to $\mathcal{H}$ because in that case either $\delta \to \gamma \to \beta \to \delta$ or $\delta \leftrightarrow \gamma \to \beta \to \delta$ and the graph would not be ancestral. Thus, $\beta \leftarrow ?\delta$ in any graph $\mathcal{G}$ joined to form $\mathcal{H}$, and hence $\beta \leftarrow ?\delta$ in $\mathcal{H}$.

**Lemma 3.3** *In a joined graph $\mathcal{H}$, formed by joining maximal ancestral graphs, if $\gamma \to \beta \to \delta? - ?\gamma$ occurs and either $\gamma \to \beta$ is real or $\beta \to \delta$ is real, then $\gamma \to \delta$ also occurs in $\mathcal{H}$. Furthermore, if both $\gamma \to \beta$ and $\beta \to \delta$ are real, then $\gamma \to \delta$ is real too.*

**Proof:** First consider the case in which the $\gamma \to \beta$ edge is real. Then, $\beta$ is not an ancestor of $\gamma$ in any $\mathcal{G}$ that gave rise to $\mathcal{H}$. If the $\{\gamma, \delta\}$ edge is undirected, or there is an arrowhead at $\gamma$ on this edge in $\mathcal{H}$, then there is some $\mathcal{G}$ that gave rise to $\mathcal{H}$ that is not ancestral. So, $\gamma \to \delta$ is in $\mathcal{H}$. A similar argument holds for the case in which the $\beta \dashrightarrow \delta$ edge is real.

If both edges $\gamma \to \beta$ and $\beta \to \delta$ are real, then $\gamma \to \delta$ also occurs in $\mathcal{H}$ and this edge is real because otherwise there is some $\mathcal{G}$ that gave rise to $\mathcal{H}$ that is not ancestral.

## 3.1 INFERRING DISCRIMINATING PATHS

The following lemma and corollary allow us to infer the presence of discriminating paths.

**Lemma 3.4** *Let $\mathcal{H}$ be a graph formed by joining a number of Markov equivalent maximal ancestral graphs. If there is a discriminating path in $\mathcal{H}$ then this discriminating path is present in every $\mathcal{G}$ joined to form $\mathcal{H}$.*

**Proof:** Suppose in some joined graph $\mathcal{H}$ there is a path $U$ as described in Definition 2.7. Label the colliders on the path between $x$ and $\beta$ as $q_1, q_2, \ldots, q_p$, such that $q_1$ is adjacent to $x$, and $q_p$ is adjacent to $\beta$. Note that $\langle x, q_1, q_2, \ldots, q_p, \beta \rangle$ forms a collider path in all $\mathcal{G}$ that gave rise to $\mathcal{H}$ because all arrowheads in $\mathcal{H}$ are also present in all $\mathcal{G}$ that gave rise to $\mathcal{H}$. Recall that $x$ and $y$ are not adjacent. There is an unshielded non-collider at $q_1$ on the path $\langle x, q_1, y \rangle$, but $x? \dashrightarrow q_1$. Because all $\mathcal{G}$ that gave rise to $\mathcal{H}$ are Markov equivalent, by Theorem 2.1 $q_1$ is a parent of $y$ in all $\mathcal{G}$ that gave rise to $\mathcal{H}$. Thus, the $\{q_1, y\}$ edge in $\mathcal{H}$ is real. We will now show by induction that all $q_m, 2 \leq m \leq p$ are also parents of $y$ in all $\mathcal{G}$ that gave rise to $\mathcal{H}$.

For $m = 2$, $\langle x, q_1, q_2, y \rangle$ discriminates $q_2$ to be a non-collider in $\mathcal{H}$. Since the $q_1 \to y$ edge is real, this discriminating path is present in all such $\mathcal{G}$, the $q_2 \to y$ edge in $\mathcal{H}$ is real. Assume for $m < p$ that $\langle x, q_1, q_2, \ldots, q_{m-1}, q_m, y \rangle$ discriminates $q_m$ to be a non-collider in all $\mathcal{G}$ that gave rise to $\mathcal{H}$ so that $q_m$ is a parent of $y$ in all $\mathcal{G}$ that gave rise to $\mathcal{H}$. Then, $U = \langle x, q_1, q_2, \ldots, q_m, q_{m+1}, y \rangle$ discriminates $\langle q_m, q_{m+1}, y \rangle$ to be a non-collider in $\mathcal{H}$. Because each of $\{q_1, q_2, \ldots, q_m\}$ is a parent of $y$ in all $\mathcal{G}$ that gave rise to $\mathcal{H}$, $U$ is a discriminating path present in all such $\mathcal{G}$ and hence the $\{q_{m+1}, y\}$ edge in $\mathcal{H}$ is real. Thus, by induction, $\langle q_1, q_2, \ldots, q_p \rangle$ are all parents of $y$ in all $\mathcal{G}$ that gave rise to $\mathcal{H}$.

But then $U^* = \{x, q_1, q_2, \ldots, q_p, \beta, y\}$ forms a discriminating path for $\beta$ in $\mathcal{H}$; $U^*$ is present in all $\mathcal{G}$ that gave rise to $\mathcal{H}$, and thus $\langle q_p, \beta, y \rangle$ forms a collider in all $\mathcal{G}$ that gave rise to $\mathcal{H}$ if and only if $\langle q_p, \beta, y \rangle$ forms a collider in $\mathcal{H}$.

**Corollary 3.1** *If a collider path $q = \langle x, q_1, \ldots, q_p, \beta \rangle$ is present in all Markov equivalent ancestral graphs* that gave rise to the joined graph $\mathcal{H}$, and $\mathcal{U} = \langle x, q_1, \ldots, q_p, \beta, y \rangle$ is a discriminating path for $\beta$ in some $\mathcal{G}$ that gave rise to $\mathcal{H}$, then $\mathcal{U}$ is also a discriminating path for $\beta$ in $\mathcal{H}$.

Just as for ancestral graphs, if a joined graph is *not* maximal, then there exists a pair of non-adjacent vertices $\{\alpha, \beta\}$ in $\mathcal{H}$ such that there is at least one inducing path between $\alpha$ and $\beta$. In the proof of the pairwise Markov property for joined graphs, we only need to consider particular inducing paths between vertices, *minimal* inducing paths.

**Definition 3.1** *Let $\mu$ be an inducing path in a joined graph with vertices $\langle \mu_0, \mu_1 \ldots, \mu_n \rangle$, and let $\psi_i$ be the number of edges on a shortest directed path between $\mu_i$ and an endpoint. Furthermore, let $\phi(\mu)$ be the total number of edges between the interior vertices and the endpoints on these paths, i.e. $\phi(\mu) = \Sigma_{i=1}^{n-1}\psi_i$. Then, $\mu$ is a "minimal inducing path" for vertices $\mu_0$ and $\mu_n$ in a joined graph if:*

*(i) There is no other inducing path between $\mu_0$ and $\mu_n$ with fewer vertices, and*
*(ii) There is no inducing path $\mu'$ with the same number of vertices as $\mu$ and $\phi(\mu') < \phi(\mu)$.*

It is easy to see that whenever there is an inducing path in a graph then there is a minimal inducing path in that graph. We use the notion of a minimal inducing path to help infer the presence and/or orientation of edges between non-consecutive vertices along a collider path in a joined graph. Lemma 3.5 and Corollary 3.2 make such inferences. See Ali and Richardson (2002) for a full proof of Lemma 3.5.

**Lemma 3.5** *Let $\mathcal{H}$ be a graph formed by joining any number of Markov equivalent maximal ancestral graphs. Suppose there is a collider path $\mu$ between vertices $\mu_0$ and $\mu_n$ such that $\mu_0$ and $\mu_n$ are not adjacent and $\mu$ is a minimal inducing path in the joined graph $\mathcal{H}$. Let $\mu_1, \mu_2, \ldots, \mu_{n-1}$ be the interior vertices along this path (i.e. the non-endpoints). If $\mu_i$ and $\mu_j$ are adjacent for $|j - i| > 1, 1 \leq i \leq n$ such that $\mu_i - ?\mu_j$ in $\mathcal{H}$, then $\mu_i \to \mu_j$ in all ancestral $\mathcal{G}$ that gave rise to $\mathcal{H}$. So, $\mu_i \to \mu_j$ occurs in $\mathcal{H}$ and this edge is real.*

Note that the $\{\mu_i, \mu_j\}$ edge cannot be bi-directed because the path $\{\mu_0, \mu_1, \ldots, \mu_i, \mu_j, \mu_{j+1}, \ldots, \mu_n\}$ is a shorter (and therefore more minimal) path than the original path $\mu$. Lemma 3.5 is useful because it tells us that whenever non-consecutive vertices along an inducing path are adjacent, these edges are directed and real. In particular, Corollary 3.2 shows that if any interior node of a collider path in a joined graph is adjacent to an endpoint then that edge is directed into the endpoint, and this edge is real.

**Corollary 3.2** *Let $\mu$ be a minimal inducing path of length $n$ in the joined graph $\mathcal{H}$. If $\mu_r, 1 \leq r \leq (n-1)$ is adjacent to an endpoint that does not occur directly before or after $\mu_r$ along the path $\mu$ (i.e. excluding $\langle \mu_0, \mu_1 \rangle$ and $\langle \mu_{n-1}, \mu_n \rangle$), then $\mu_r$ is a parent of the endpoint in $\mathcal{H}$ and this edge is real.*

**Proof:** If the endpoint is a parent or spouse of $\mu_r$ then the minimality of the path is violated. By Lemma 3.5 there must be a directed edge between $\mu_r$ and the endpoint. Therefore $\mu_r$ is a parent of the endpoint, and by Lemma 3.5, this edge is real.

## 4 MAXIMALITY OF JOINED GRAPHS

To prove that joining Markov equivalent maximal ancestral graphs results in a maximal joined graph, we show that if there is a minimal inducing path, $\mu$, in the joined graph, with endpoints $x$ and $y$, then $\mu$ is present in all the ancestral graphs $\mathcal{G}$ that gave rise to the joined graph.

Since the graphs that are joined are assumed to be maximal this implies that $x$ and $y$ are adjacent in all $\mathcal{G}$, and hence in $\mathcal{H}$. Recall that inducing paths are collider paths such that each interior node is an ancestor of at least one endpoint. Consider any interior node, $\mu_i, 0 < i < n$ where $n$ is the length of the path $\mu$, and let $\alpha_0$ be the first descendant of $\mu_i$ on the directed path from $\mu_i$ to an endpoint. The crux of the proof lies in showing that $\mu_i$ is a parent of $\alpha_0$ in all ancestral graphs $\mathcal{G}$ that gave rise to the joined graph.

We first prove the case in which $n = 2$, and then state the case for which $n > 2$. A full proof of Lemma 4.2 is given in Ali and Richardson (2002).

**Lemma 4.1** *Let $\mathcal{H}$ be a graph formed by joining a number of Markov equivalent maximal ancestral graphs $\mathcal{G}$. Suppose the shortest minimal inducing path with non-adjacent endpoints in $\mathcal{H}$ is of length 2 and label the nodes $\{\mu_0, \mu_1, \mu_2\}$. Call this path $\mu$. Let $\alpha_0$ be the first descendant of $\mu_1$ on the ancestral path from $\mu_1$ to an endpoint in $\mathcal{H}$. Then the $\mu_1 \to \alpha_0$ edge in $\mathcal{H}$ is real.*

**Proof:** Suppose for a contradiction that the $\mu_1 \to \alpha_0$ edge in $\mathcal{H}$ is not real. Then there is at least one ancestral graph $\mathcal{G}$ that gave rise to $\mathcal{H}$ in which $\mu_1 \leftrightarrow \alpha_0$ occurs. Since all graphs that gave rise to $\mathcal{H}$ are Markov equivalent, by Theorem 2.1 the $\{\mu_1, \alpha_0\}$ edge is shielded from both sides; i.e. $\mu_0$ and $\mu_2$ are each adjacent to $\alpha_0$: if $\mu_i, i = 0, 2$ is not adjacent to $\alpha_0$ then $\langle \mu_i, \mu_1, \alpha_0 \rangle$ forms an unshielded non-collider that is present in all Markov equivalent $\mathcal{G}$ that gave rise to $\mathcal{H}$.

There is at least one ancestral $\mathcal{G}$ that gave rise to $\mathcal{H}$ such that $\mu_1 \to \alpha_0$. In this $\mathcal{G}$, where $\mu_1$ is a parent of $\alpha_0$, $\mu_0$ and $\mu_2$ are either parents or spouses of $\alpha_0$ because otherwise either $\mu_0? \to \mu_1 \to \alpha_0 \to \mu_0$ or $\mu_2 \leftarrow \alpha_0 \leftarrow \mu_1 \leftarrow?\mu_2$ would form a non-ancestral configuration (see Figure 6). In other words, whenever $\mu_1 \to \alpha_0$ occurs in some $\mathcal{G}$, $\mu_0? \to \alpha_0 \leftarrow?\mu_2$ also occurs. Note that $\mu_0? \to \alpha_0 \leftarrow?\mu_2$ cannot occur in $\mathcal{H}$ because if it did, then the minimality of the path $\mu$ would be violated. Since $\alpha_0$ is sometimes a collider and sometimes a non-collider on the path $\{\mu_0, \alpha_0, \mu_2\}$ in the ancestral graphs that gave rise to $\mathcal{H}$, $\mu_0$ and $\mu_2$ are adjacent by Theorem 2.1. But this is a contradiction. Therefore, the $\mu_1 \to \alpha_0$ edge is real.

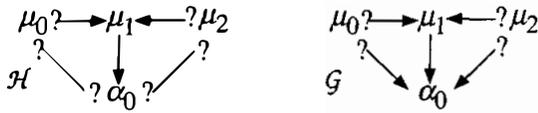

**Figure 6:** If $n = 2$ and the path $\mu$ is as outlined in Lemma 4.1, then $\mu_0$ and $\mu_2$ are adjacent to $\alpha_0$.

**Lemma 4.2** *Let $\mathcal{H}$ be a graph formed by joining a number of Markov equivalent maximal ancestral graphs. Consider the shortest minimal inducing path in $\mathcal{H}$ with non-adjacent endpoints; label the nodes $\{\mu_0, \mu_1, \ldots, \mu_{n-1}, \mu_n\}$, and call this path $\mu$. Let $\alpha_0$ be the first descendant of $\mu_i$ on the shortest ancestral path from $\mu_i$ to an endpoint, $0 < i < n$. Then $\alpha_0$ is a child of $\mu_i$ in all ancestral graphs $\mathcal{G}$ that gave rise to $\mathcal{H}$ (i.e. $\mu_i \to \alpha_0$ is real).*

We now prove the main result of this paper:

**Theorem 4.1** *Let $\mathcal{H}$ be a graph formed by joining a number of Markov equivalent ancestral graphs $\mathcal{G}$: If all $\mathcal{G}$ are maximal, then $\mathcal{H}$ is also maximal.*

**Proof:** For a contradiction, let us assume that $\mathcal{H}$ is not maximal. Then, there must be at least one pair of non-adjacent vertices such that there is an inducing path between them. Consider the shortest minimal inducing path with non-adjacent endpoints in $\mathcal{H}$; label the nodes $\{\mu_0, \mu_1, \ldots, \mu_{n-1}, \mu_n\}$, and call this path $\mu$.

From the definition of an inducing path, we know that $\mu$ is a collider path such that $\{\mu_1, \ldots, \mu_{n-1}\}$ are each ancestors of either $\mu_0$ or $\mu_n$ in $\mathcal{H}$. By the minimality of the path $\mu$, we also know that there is no inducing path on a subset of $\mu$ for which the endpoints are not adjacent. Finally note that the collider path between $\mu_0$ and $\mu_n$ is also present in all ancestral graphs, $\mathcal{G}$, that gave rise to $\mathcal{H}$.

We will show that all the edges on the directed path from $\mu_i$ to an endpoint are real. Hence, we will have shown that the inducing path in $\mathcal{H}$ is also present in all $\mathcal{G}$ that gave rise to $\mathcal{H}$. Since all $\mathcal{G}$s that gave rise to $\mathcal{H}$ are maximal, if $\mu$ is an inducing path in all $\mathcal{G}$, then $\mu_0$ and $\mu_n$ are adjacent in all $\mathcal{G}$; hence they are adjacent in $\mathcal{H}$, and we reach a contradiction.

Let $\mu_i \to \alpha_0 \to \ldots \to \alpha_m$ be the shortest directed path from $\mu_i$ to an endpoint in $\mathcal{H}$ (so $\alpha_m = \mu_0$ or $\alpha_m = \mu_n$). It is sufficient to show that this path is present in any $\mathcal{G}$ joined to form $\mathcal{H}$ (because in that case there would be at least one ancestral graph $\mathcal{G}$ that gave rise to $\mathcal{H}$ that was not maximal, which would be a contradiction).

Lemma 4.2 shows that the first descendant, $\alpha_0$, of any vertex along $\mu$, say $\mu_i$, is a child of $\mu_i$ in all $\mathcal{G}$ that gave rise to $\mathcal{H}$ (i.e. $\mu_i \to \alpha_0$ is real). We will use an inductive proof to show that all the subsequent edges on the directed path from $\mu_i$ to an endpoint are also real.

Suppose that $\alpha_0 \to \alpha_1$ is not real. Then by Theorem 2.1, $\mu_i \to \alpha_0 \to \alpha_1$ is shielded (i.e. $\mu_i$ is adjacent to $\alpha_1$) since otherwise $\langle \mu_i, \alpha_0, \alpha_1 \rangle$ is an unshielded collider. By Lemma 3.3 $\mu_i \to \alpha_1$ is in $\mathcal{H}$ because the $\mu_i \to \alpha_0$ edge is real; but then the minimality of the path is violated since we can take a directed path from $\mu_i$ to an endpoint that bypasses $\alpha_0$ (see Figure 7). So, the edge $\alpha_0 \to \alpha_1$ is real.

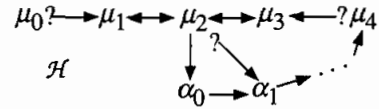

**Figure 7:** If the path $\mu$ is as outlined in Lemma 4.1, then $\mu_2$ and $\alpha_1$ are adjacent, and $\mu_2$ is a parent of $\alpha_1$.

Assume that for $0 < k < m$, the $\alpha_{k-1} \to \alpha_k$ edge is real and consider the $\alpha_k \to \alpha_{k+1}$ edge. If $\alpha_k \to \alpha_{k+1}$ in $\mathcal{H}$ is not real, then by Theorem 2.1, $\alpha_{k-1}$ is adjacent to $\alpha_{k+1}$. By Lemma 3.3, $\alpha_{k-1} \to \alpha_{k+1}$ occurs in $\mathcal{H}$ (because the $\alpha_{k-1} \to \alpha_k$ edge is real), but then the minimality of the path is violated since we can take the path from $\mu_i$ to an endpoint that bypasses $\alpha_k$. Consequently, the $\alpha_k \to \alpha_{k+1}$ edge is real. So, by induction, the edges between $\alpha_0, \alpha_1, \alpha_2, \ldots, \alpha_m$ are real and form a directed path from $\alpha_0$ to an endpoint.

We have now shown that $\mu$ also occurs in all $\mathcal{G}$ that gave rise to $\mathcal{H}$. Since all these ancestral graphs $\mathcal{G}$ are maximal, $\mu_0$ and $\mu_n$ are adjacent in every $\mathcal{G}$. Consequently, $\mu_0$ and $\mu_n$ are adjacent on $\mathcal{H}$ which is a contradiction.

## 5 Conclusions and Future Work

Ancestral graphs are a class of graphs that can represent the independence relations holding among the observed variables of a DAG model with latent and selection variables. Unfortunately, as with DAG models, there often are a number of ancestral graphs that can encode the same independence model. We have defined a new, broader class of graphs, joined graphs, which extract the arrowheads common to Markov equivalent ancestral graphs. Ancestral graphs (and therefore directed acyclic graphs) form a subset of joined graphs. The goal of introducing joined graphs is to associate a unique graph with each equivalence class of ancestral graphs.

A full characterization of equivalence classes for ancestral graphs is our ultimate goal. In this paper we have taken a step towards this by proving the pairwise Markov property for joined graphs. The next step will be to prove the global Markov property for joined graphs. However, there are a number of other interesting questions that have yet to be answered: For which graphs $\mathcal{H}$, does there exist a set of Markov equivalent maximal ancestral graphs that can be joined to give rise to $\mathcal{H}$? Is there a set of orientation rules such that given a member of an equivalence class, one could construct the corresponding joined graph for the entire equivalence class? If an equivalence class contains a DAG, is the joined graph for the entire class the same as the essential graph for the equivalence class for DAGs? The authors are in the process of investigating these issues.

### Acknowledgments

We would like to acknowledge Michael Perlman and the reviewers for their valuable comments on previous drafts of this paper. This research was supported by the U.S. National Science Foundation under grant DMS-9972008, and by the U.S. Environmental Protection Agency.